\pdfoutput=1

\documentclass[11pt]{article}

\usepackage{EMNLP2022}

\usepackage{times}
\usepackage{booktabs}
\usepackage{multirow}
\usepackage{latexsym}
\usepackage{graphicx,booktabs,array}
\usepackage{subfigure} 
\usepackage[T1]{fontenc}
\usepackage{geometry}

\usepackage[utf8]{inputenc}

\usepackage{microtype}

\usepackage{inconsolata}

\title{TRIPS: Efficient Vision-and-Language Pre-training with Text-Relevant Image Patch Selection}

\author{{Chaoya Jiang$^{1}$, Haiyang Xu $^2$, Chenliang Li $^2$, Ming Yan $^2$ } \\ 
\textbf{Wei Ye $^1$\thanks{ ~corresponding author.} , Shikun Zhang $^{1}$\footnotemark[1], Bin Bi $^2$, Songfang Huang $^2$} \\
$^1$ National Engineering Research Center for Software Engineering, Peking University \\
$^2$ DAMO Academy, Alibaba Group \\
\{jiangchaoya,wye, zhangsk\}@pku.edu.cn,\\
\{shuofeng.xhy,ym119608,lcl193798,b.bi, songfang.hsf\}@alibaba-inc.com
}

\begin{document}
\maketitle
\begin{abstract}

Vision Transformers (ViTs) have been widely used in large-scale Vision and Language Pre-training (VLP) models. Though previous VLP works have proved the effectiveness of ViTs, they still suffer from computational efficiency brought by the long visual sequence. To tackle this problem, in this paper, we propose an efficient vision-and-language pre-training model with \textbf{T}ext-\textbf{R}elevant \textbf{I}mage \textbf{P}atch \textbf{S}election, namely TRIPS, 
which reduces the visual sequence progressively with a text-guided patch-selection layer in the visual backbone for efficient training and inference. The patch-selection layer can dynamically compute text-dependent visual attention to identify the attentive image tokens with text guidance and fuse inattentive ones in an end-to-end manner. Meanwhile, TRIPS does not introduce extra parameters to ViTs. Experimental results on a variety of popular benchmark datasets demonstrate that TRIPS gain a speedup of 40\% over previous similar VLP models, yet with competitive or better downstream task performance.

\end{abstract}
\section{Introduction}
In recent years, Vision-Language Pre-training (VLP)~\cite{Tan2019LXMERTLC, Chen2019UNITERLU,Lu2019ViLBERTPT,Huang2020PixelBERTAI,Su2020VLBERTPO, Li2020OscarOA,Chen2020UNITERUI,Zhou2020UnifiedVP,Li2021AlignBF,Yu2021ERNIEViLKE,li2022mplug} has developed at an astonishing rate and become a prevalent paradigm to tackle VL tasks. Traditional VLP models~\cite{Tan2019LXMERTLC, Chen2019UNITERLU,Lu2019ViLBERTPT,Li2020OscarOA} utilize pre-trained object detectors \cite{Ren2015FasterRT,Redmon2016YouOL,He2017MaskR} to extract region-based image features but suffer from extensive annotation and expensive computation of object detector training. 
Inspired by the success of the Vision Transformer (ViT)~\cite{Dosovitskiy2021AnII} and its variants~\cite{Liu2021SwinTH,Wu2021CvTIC,Wang2021PyramidVT} in computer vision field, more recent VLP models ~\cite{Li2021AlignBF,Radford2021LearningTV,Kim2021ViLTVT,Wang2021VLMoUV,Singh2021FLAVAAF} have adopted ViT as the visual encoder or cross-modal fusion encoder without using region features from the pre-trained object detector.


\begin{figure*}[h]
\centering


\centering
\includegraphics[width=0.98\linewidth]{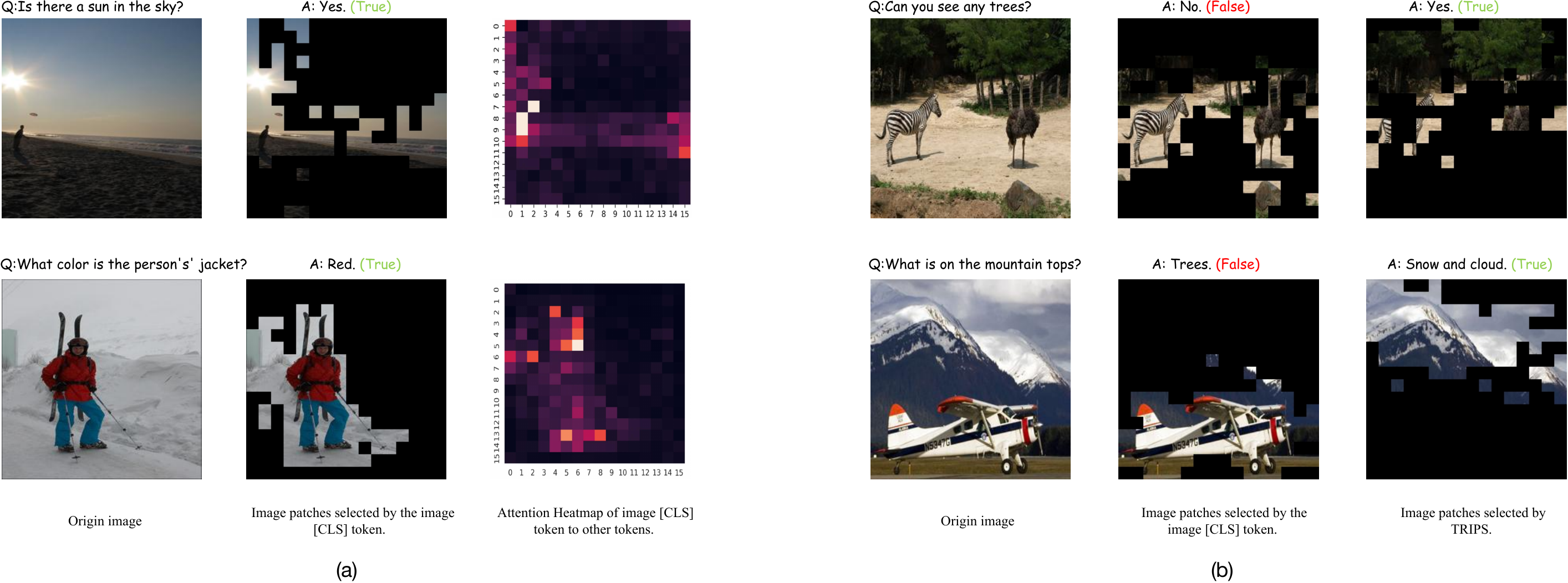}
\caption{The sub-figure (a) shows the VQA cases of the ALBEF~\cite{Li2021AlignBF} finetuned on the VQA task, in which the input image tokens are directly selected with the guidance of attention weight of the image [CLS] token to other image tokens. We visualize the attention distribution of the image [CLS] token. As we can see, the [CLS] token attention naturally focuses on the objects in images and ignores the image backgrounds. If the question is about the objects in the images, we can still get the correct prediction. The sub-figure (b) shows the VQA predictions comparison between ALBEF, which directly selects image tokens with the guidance of the image [CLS] token, and our model TRIPS. As we can see, the questions are about the image backgrounds, and the former predicts the wrong answers, yet the latter can give the correct answer as it preserves the text-relevant image tokens. }
\label{fig1}

\end{figure*}

However, these ViT-based VLP methods are required to model long visual sequences from high-resolution images for good vision understanding, with quadratic computational complexity to the length of the visual sequence. Moreover, recent efforts \cite{Dosovitskiy2021AnII, Touvron2021TrainingDI} have also begun to explore vision-language foundation models, which scale up the model and data size. This raises the necessity to decrease the high computational cost of ViT-based VLP models.
As shown in Figure \ref{fig1}(a), we find that removing the inattentive patch tokens of the image [CLS] token will generally not affect the result of Visual Question Answering (VQA)~\cite{Agrawal2015VQAVQ} prediction. Based on the observation, we conjecture that not all image tokens in the visual encoder contribute positively to the final prediction result of VLP models, and large numbers of redundant image tokens exist. 

There have been some recent studies ~\cite{Rao2021DynamicViTEV,Xu2021EvoViTST,Zong2021SelfslimmedVT, Liang2022NotAP} focusing on ViT model accelerations by reducing unrelated image tokens. However, these methods are specially designed for computer vision tasks (e.g., image recognition) and remove the redundant tokens based on visual semantics,  ignoring the aligned knowledge in text modality, and thus are not suitable for VL tasks. As shown in Figure \ref{fig1}(b), without the guidance of text knowledge, directly removing patch tokens based on the image [CLS] token will lead to a wrong answer. This observation motivates us to reduce the image tokens by fusing the less informative patch tokens with the guidance of aligned text context.

In this work, we propose an efficient VLP model with Text-Relevant Image Patch Selection (TRIPS), to progressively reduce the redundant image tokens with text guidance. TRIPS selects text-consistent image tokens through the text-aware patch-selection layer, reducing the computational cost of the visual encoding and cross-modal fusion. The patch-selection layer can preserve the attentive image tokens with text guidance and fuse the inattentive tokens into a single one by dynamically computing text-dependent visual attention, in an end-to-end way. In this way, we gradually decrease the number of image tokens as the visual backbone goes deeper to alleviate the computational cost of the visual encoder while improving the efficiency of cross-modal fusion due to the reduction of visual sequences. Besides, the model efficiency can be flexibly controlled via the keep rate of image tokens in the patch-selection layer where no additional parameters are introduced. We evaluate TRIPS on various representative VL tasks, including visual question answering, natural language visual reasoning, and cross-modal retrieval. The efficiency of TRIPS can be increased by 40\% compared to previous similar VLP models, yet with competitive or better downstream task performance. For instance,  TRIPS can speed up the baseline model 40.98\% and even improve by 0.1 on the VQA test-dev and 0.2 on the NLVR Dev with the same experimental settings (see Table \ref{table3}). Furthermore, by increasing the input image resolution with the same computational cost, TRIPS can improve 0.4 on the VQA test-dev and 0.6 on the NLVR Dev.
Our contributions can be summarized as three-fold:

\begin{itemize}

\item We propose an efficient vision-and-language pre-training model with Text-Relevant Image Patch Selection (TRIPS). As far as we know, this is the first exploration that decreases the computational cost of VLP models by reducing image tokens with the help of linguistic context.
\item We propose a text-relevant patch-selection layer, which can dynamically compute text-dependent visual attention to identify the attentive image tokens and fuse inattentive ones with text guidance in an end-to-end manner.
\item Extensive experiments indicate that our TRIPS can boost the VLP model efficiency at a lower computational cost than the un-accelerated baseline model. Furthermore, by increasing the input image resolution, TRIPS benefits from taking more image tokens to achieve better performance without increasing computational costs.
\end{itemize}

\section{Related Work}
\subsection{Vision-Language pre-training}
Previous Vision-Language pre-training (VLP) methods \cite{Lu2019ViLBERTPT,Li2019VisualBERTAS, Tan2019LXMERTLC, Li2020OscarOA,Chen2019UNITERLU,Yu2021ERNIEViLKE} mainly take a two-step training pipeline approach, which first extracts visual features by a pre-trained object detector and then trains the cross-modal pre-training model to align text and visual features. Some region-based methods reduce the computation cost with the lightweight model architecture \cite{wang2020minivlm,gan2021playing} and knowledge distillation \cite{fang2021compressing}. 
The main challenge for these methods is to balance effectiveness and efficiency. More recent CNN-based \cite{Huang2020PixelBERTAI, Xu2021E2EVLPEV} and ViTs-based \cite{li2021align,Kim2021ViLTVT,Radford2021LearningTV,Wang2021VLMoUV} methods (especially the patch-based ViT) removes the complicated object detector in feature extraction to conductsnd-to-end VL learning. However, there is no work decreasing the high computational cost of ViT-based VLP models. In this work, we propose a novel method to decrease the computation cost of VLP models, which reduces the visual sequence progressively with a text-guided patch-selection layer in the visual backbone for efficient training and inference.

\subsection{ViTs Acceleration}
To accelerate the computation of the transformer\cite{Vaswani2017AttentionIA} based model, many studies focus on proposing more efficient attention mechanisms \cite{Wang2020LinformerSW, Kitaev2020ReformerTE, Choromanski2021RethinkingAW} or compress Transformer structures \cite{Liu2021SwinTH, Heo2021RethinkingSD,Wang2021PyramidVT}. Recently, some approaches have focused on accelerating ViTs by reducing the number of tokens involved in the inference of ViTs. For example, to expedite ViTs, \citet{Ryoo2021TokenLearnerAS} proposed TokenLearner, in which a relatively small amount of tokens are learned by aggregating the entire feature map weighted by dynamic attention. \citet{Rao2021DynamicViTEV} introduces a method to reduce tokens for a fully trained ViT, where an extra learnable neural network is added to ViT to select a subset of tokens. \citet{Liang2022NotAP} propose to reduce the computational overhead of inference by proposing a token reorganization method to reduce and reorganize image tokens progressively. However, those methods are unsuitable for VLP as they reduce the image tokens without considering the text context. 

 \begin{figure*}[htpb]
\setlength{\belowcaptionskip}{-0.15cm}
\centering
\includegraphics[width=0.97\textwidth]{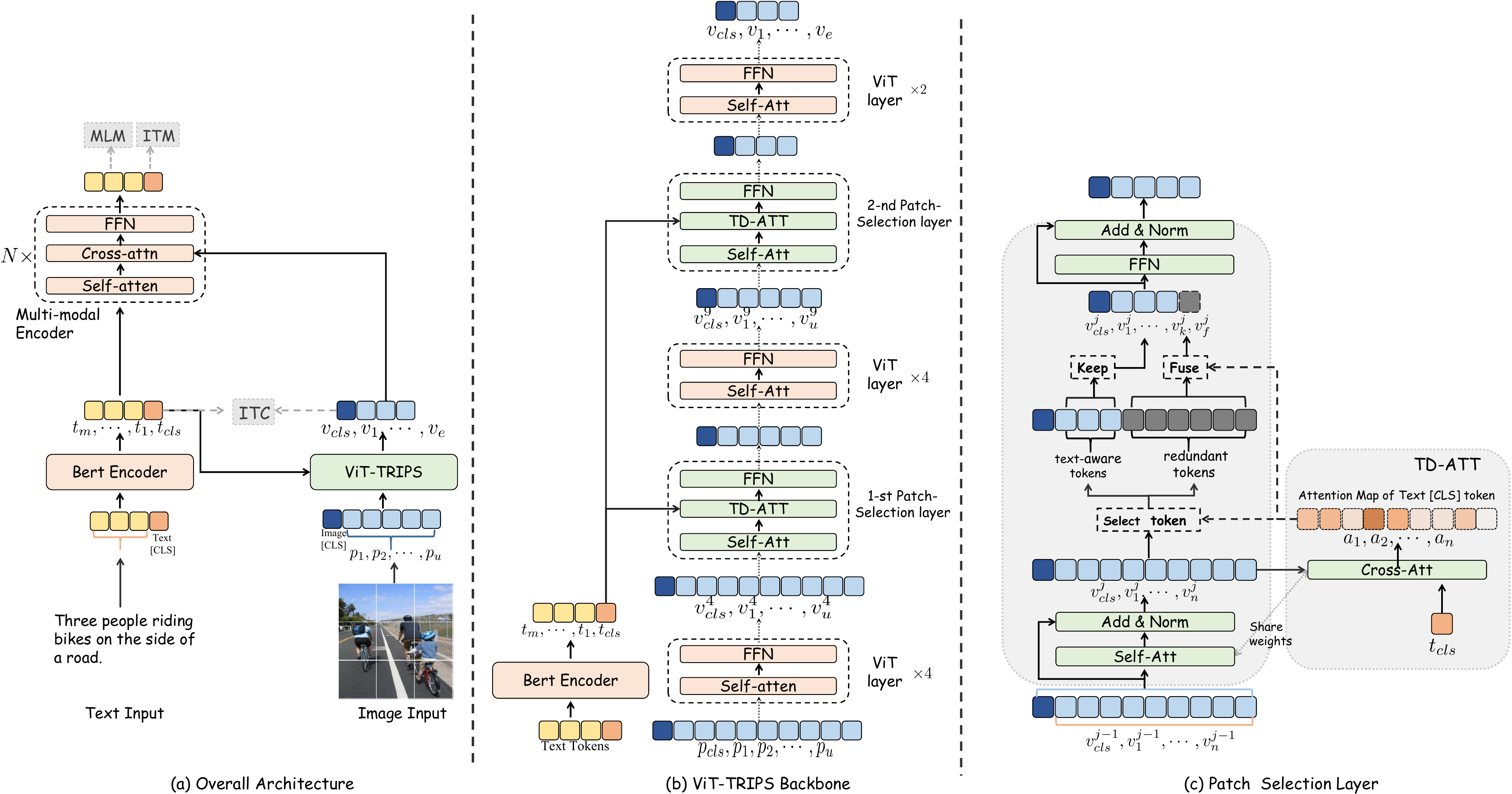}

\caption{(a) The overall architecture of our VLP model (TRIPS) in this paper. (b) An overview of the ViT applied with a Text-Relevant Image Patch Selection module (ViT-TRIPS). Suppose the ViT-TRIPS contain 12 layers, with the 5-th and 10-th layer as the patch selection layers.  (c)  Illustration of Text-Relevant Image Patch-Selection layer.}

\label{fig2}
\end{figure*}

\section{Method}

In this section, we will first introduce TRIPS with the acceleration module of the text-relevant patch-selection layer, and then give the details of the pre-training objectives.

\subsection{Model Architecture}

As shown in Figure~\ref{fig2} (a), TRIPS contains a visual encoder with text-relevant patch-selection layer, a text encoder, and a multimodal fusion encoder. The visual encoder takes a  Vision Transformer (ViT), where text-relevant patch-selection layers are used to progressively reduce and reorganize image tokens, namely ViT-TRIPS. The text encoder adopts BERT$_{base}$ transformer \cite{Devlin2019BERTPO}. Similar with~\citep{Li2021AlignBF}, the multimodal fusion encoder is a transformer encoder that performs the cross-modal interaction and fusion through a cross-attention mechanism. 

Formally, given an input image-text pair, we first feed the input text to the text encoder and represent it as a sequence of embeddings $T = \{t_{cls}, t_1, t_2, \cdots, t_m \}$, where $t_{cls}$ is the embedding of the text [CLS] token to summarize the input text. Then, we divide the input image into patches $P = \{p_{cls}, p_1,p_2, \cdots, p_u \}$, and encode them with the image encoder ViT-TRIPS. It takes the text [CLS] embedding $t_{cls}$ and image patches $\{p_{cls}, p_1,p_2, \cdots, p_u \}$ as input, and outputs the image sequence $V=\{v_{cls}, v_1, v_2, \cdots,v_e\}$. Note that $e < u$, since we apply the text-relevant patch-selection layer to select the text-aware image tokens and fuse the redundant tokens, allowing us to reduce total visual sequence length for efficiency. Finally, the text features $ \{t_{cls}, t_1, t_2, \cdots, t_m \}$  and the image features  $\{v_{cls}, v_1,v_2, \cdots, v_e \}$ encoded by the image encoder are fused by cross attention at each layer of the multimodal encoder as in ALBEF~\cite{Li2021AlignBF}. The output of the multimodal encoder is used to pre-train and finetune downstream tasks. 


\noindent \subsection{Text-Relevant Image Patch Selection}\label{sect2}
Existing works in computer vision~\cite{Rao2021DynamicViTEV, Liang2022NotAP} select patches by using only the image [CLS] token from the ViT backbone. However, as shown in Figure \ref{fig1} (b), the selection of image tokens in cross-modal tasks is closely related to textual context, and different texts for a single image may focus on different parts of the visual content. Selecting the image tokens with the guidance of aligned textual content can help the VLP model focus on the key parts of the image for more effective and efficient cross-modal fusion. Here, we present a text-relevant patch-selection layer that can dynamically select the image patches with the guidance of textual input, yet with no additional parameters introduced.

As shown in Figure \ref{fig2} (b), for a ViT with $L$ standard Transformer layers and $t$ patch-selection layers in total, the interval length is obtained as $s = L / (t + 1)$. Then, we choose the layer index $j=i*s+1$ as the $i_{th}$ patch-selection layer, so that patch-selection layers are uniformly inserted into the ViT-TRIPS backbone. In each patch-selection layer, as shown in Figure \ref{fig2}(c), we adopt standard self-attention (SA), Text-aware Dynamic Attention (TD-ATT), and Inattentive Token Fusion (ITF) modules to progressively reduce image tokens.

Specifically, for the $i_{th}$ patch-selection layer, image features $v^{j-1}=\{v^{j-1}_{cls}, v^{j-1}_1, \cdots, v^{j-1}_n\}$ are first fed to the $j_{th}$ visual self-attention layer:

\begin{equation}
v^{j}=LN(SA(v^{j-1})+v^{j-1})
\end{equation}
where LN is short for layer normalization, and $n$ is the number of patch tokens in $j-1$ visual transformer layer. 
 
Next, we will illustrate how to use the Text-aware Dynamic Attention mechanism (TD-ATT) to select the text-aware image patch tokens. The text [CLS] embedding $t_{cls}$ is linearly projected to the query vector denoted as $q_{text}$ by the shared query linear layer of the $j_{th}$ visual self-attention layer. We compute the text-to-image attention feature map excluding the image [CLS] token as follows:
\begin{equation}
    a_{cls} = softmax(\frac{ q_{text} \cdot v^{j}[1:]^T}{\sqrt{d}})
\end{equation}

We identify and preserve the attentive image tokens
corresponding to the $k$ largest elements in the attention map $a_{cls}=\{a_{1}, ..a_{n} \}$, where $k = n \times r$, and $r$ is the keep rate of this layer. 
The selected image tokens are kept and the un-selected image tokens are further fused by an inattentive token fusion operation ITF.

The remaining inattentive patch tokens $\{v_{z_1}, v_{z_2}, \cdots,v_{z_{n-k}}\}$ are treated as text-irrelevant tokens. However, the fixed keep rate may remove some useful tokens, so we fuse inattentive tokens to one token $v_{f}$ by a weighted sum operation to supplement attentive ones as follow:
\begin{equation}
    v_{f} = \sum\limits^{n-k}\limits_{i = 1} a_{cls,z_i} \cdot \hat{v}_{z_i}
\end{equation}
After fusing the inattentive patch tokens,  we reconstruct the $j_{th}$ visual sequence as $v^{j} = \left[v^{j}_{cls},v^{j}_{1}, \cdots,v^{j}_{k},{v}^{j}_{f}\right]$, which consists of the image [CLS] token embedding, the selected text-aware image patch embedding, and fused inattentive patch embedding. Then the new visual sequence is fed to the feed-forward network (FFN). 


\paragraph{Extension to Single-stream Model.} The proposed Text-Relevant Image Patch Selection layer can be also extended to the single-stream model (TRIPS-S), which employs the [CLS] token of the multimodal encoder to preserve the attentive image tokens and fuses the inattentive tokens to speed up the training and inference. While parameter-efficient, it may be difficult to learn uni-modal and multi-modal interactions concurrently. Therefore, its performance lags behind two-stream performance on downstream VL tasks.

\subsection{Pre-training Objectives}

We pre-train our model with three standard objectives: Image-Text Contrastive learning (ITC), Image-Text Matching (ITM), and  Masked Language Modeling (MLM). These pre-training tasks are optimized jointly. 

\noindent \textbf{Image-text Contrastive (ITC)} For TRIPS, We follow the \cite{Li2021AlignBF} and apply ITC to align the image representation and text representation from the unimodal encoders. For the image, the image feature corresponding to the image [CLS] token is chosen as the image representation. For the text, the text token feature corresponding to the text [CLS] token is the text representation.

\noindent \textbf{Image-Text Matching (ITM)}  The goal of image-text matching is to predict whether the input image and text are matched.  We follow the design of \cite{Li2021AlignBF} and select hard negative image-text pairs based on the contrastive text-image similarity. We take the text [CLS] embedding of the multimodal encoder's output as the joint representation, followed by a Multi-Layer Perceptron (MLP) layer for prediction.

\noindent \textbf{Masked Language Modeling (MLM)} The task setup is basically the same as in BERT~\cite{Devlin2019BERTPO}, where we randomly mask 15$\%$ of tokens in text and the model is asked to predict these masked words with the cross-modal representations.

\begin{table*}[h]\normalsize
\setlength{\tabcolsep}{1.4mm}{
\begin{tabular}{c|cccccc|cccccc}
\toprule[2.0pt]
\multirow{3}{*}{Models}   & \multicolumn{6}{c|}{Flickr30K (1K test set)} & \multicolumn{6}{c}{MSCOCO (5K test set)} \\
            &\multicolumn{3}{c}{TR}                                                         & \multicolumn{3}{c|}{IR}                                                         & \multicolumn{3}{c}{TR}                                                         & \multicolumn{3}{c}{IR}                                                         \\
                    & R@1                      & R@5                      & R@10                     & R@1                      & R@5                      & R@10                      & R@1                      & R@5                      & R@10                     & R@1                      & R@5                      & R@10                     \\ \hline
E2E-VLP          & 86.2                     & 97.5                     & 98.92                    & 73.6                     & 92.4                     & 96.0                      & -                        & -                        & -                        & -                        & -                        & -                        \\
UNITER        & 87.3                     & 98.0                     & 99.2                     & 75.6                     & 94.1                     & 96.8                      & 65.7                     & 88.6                     & 93.8                     & 52.9                     & 79.9                     & 88.0                     \\
OSCAR         & -                        & -                        & -                        & -                        & -                        & -                         & 70.0                     & 91.1                     & 95.5                     & 54.0                     & 80.8                     & 88.5                     \\
VinVL         & -                   & -                    & -                     & -& - & -                  &              74.6            & 92.6                       & 96.3                       & 58.1                        & 83.2                        & 90.1                        \\
ViLT       & \multicolumn{1}{c}{83.5} & \multicolumn{1}{c}{96.7} & \multicolumn{1}{c}{98.6} & \multicolumn{1}{c}{64.4} & \multicolumn{1}{c}{88.7} & \multicolumn{1}{c|}{93.8} & \multicolumn{1}{c}{61.5} & \multicolumn{1}{c}{86.3} & \multicolumn{1}{c}{92.7} & \multicolumn{1}{c}{42.7} & \multicolumn{1}{c}{72.9} & \multicolumn{1}{c}{83.1} \\
ALBEF       & 94.3                     & 99.4                     & 99.8                     & 82.8                     & 96.7                     & 98.4                      & 73.1                     & 91.4                     & 96.0                     & 56.8                     & 81.5                     & 89.2                     \\\hline
  ALBEF-C &  96.1   &   99.8    &   \textbf{100.0}    &    \textbf{86.1}   &    97.8   &   98.9 & 77.8    &    94.3   &    97.4   &   60.3    &   \textbf{ 84.7}   &   91.0       \\

TRIPS   &     \textbf{96.3}                    &               \textbf{ 99.8}          &      \textbf{100.0}                   &         85.8                 &      \textbf{98.1}                    &           \textbf{99.0}                &       \textbf{78.1}                   &        \textbf{ 94.8}                 &     \textbf{97.6}                     &       \textbf{ 61.3}                  &                 84.3         &           \textbf{91.4}               \\ \bottomrule[2.0pt]
\end{tabular}}
 \caption{Image-text retrieval results on Flickr30K and COCO datasets. $-$indicate that the result is unreported.}.
\centering
\label{table2}
\end{table*}

\begin{table}[h]
\setlength{\tabcolsep}{1.6mm}
\begin{tabular}{ccccc}
\toprule[2.0pt]
\multirow{2}{*}{Models} & \multicolumn{2}{c}{VQA} & \multicolumn{2}{c}{NLVR} \\
                                    & Test-dev   & Test-std   & dev        & Test-P      \\ \hline
ViLBERT                           & 70.55      & -          & -      & -       \\
LXMER                        & 72.42      & -          & 74.90      & 74.50       \\
UNITER                          & 72.70      & 72.91      & 77.18      & 77.85       \\
OSCAR              & 73.16      & 73.44      & 78.07      & 78.36       \\
VinVL                & 75.95      & 76.12          & 82.05       & 83.08       \\
E2E-VLP       & 73.25     & 73.67          & 77.25       & 77.96       \\
ViLT                            & 71.26      & -          & 75.70       & 76.13       \\
ALBEF                          & 74.54      & 74.70      & 80.24      & 80.50        \\ \hline
ALBEF-C          &  76.12     &76.32    &  82.21          &   83.11                   \\
TRIPS                          &  \textbf{76.23}          &   \textbf{76.48}         &      \textbf{82.35}      &     \textbf{83.34}       \\ \bottomrule[2.0pt]
\end{tabular}
\caption{Evaluation Results on VQA test set and NLVR2. $-$ indicates that the result is unreported. ALBEF-C is our implementation of ALBEF with a visual encoder initialized by CLIP (ViT-B/16). TRIPS takes the same model architecture and experimental setting as ALBEF-C.}
\label{table1}

\end{table}

\section{Experiment Settings}
\subsection{Implementation Details}

We pre-train the TRIPS for 30 epochs with a total batch size of 512 on 8 NVIDIA V100 GPUs. We initialize the visual encoder by CLIP (ViT-B/16) \cite{Radford2021LearningTV} pre-trained on 400M noisy image-text pairs and we use the AdamW \cite{Loshchilov2019DecoupledWD} optimizer with a weight decay of 1e-2. The learning rate is warmed-up to 1e-5 (ViT-B$/$16) and 1e-4 (BERT$_{base}$) in the first 1000 iterations.  During pre-training, we take the image with the resolution of $256 \times 256$ as input and increase the image resolution during finetuning. We use a 6-layer Transformer for both the text encoder and the cross-modal fusion network. As ~\citep{Li2021AlignBF}, the text encoder is initialized using the first 6 layers of the  BERT$_{base}$~\cite{Devlin2019BERTPO} model and the cross-modal network is initialized using the last 6 layers of the~BERT$_{base}$. For the image-text contrastive learning, the queue size is set as 65,536 and the momentum coefficient is set as 0.995.  

\subsection{Pre-training Datasets}
We construct our pre-training data using two web datasets (Conceptual Captions \cite{Sharma2018ConceptualCA}, SBU Captions \cite{Ordonez2011Im2TextDI}) and two in-domain datasets (MSCOCO \cite{Lin2014MicrosoftCC} and Visual Genome \cite{Krishna2016VisualGC}). The total number of unique images is 4.0M, and the number of image-text pairs is 5.1M.

\begin{table}[h]
\setlength{\tabcolsep}{2.1mm}
\begin{tabular}{ccccc}
\toprule[2.0pt]
Models  & FLOPs & Throughput & Latency \\ \hline
UNITER    &    949.90    &     6.42       &     ~870ms    \\
OSCAR     &    956.40  &    6.35      &     ~860ms    \\
VinVL      &    1023.30    &    7.32         &   ~640ms     \\
E2E-VLP    &    144.3     &     80.23     &     ~70ms    \\
ViLT     &    55.40   & 247.530       &  ~15ms    \\
ALBEF-C   &    36.63    &     197.52    &      ~21ms   \\
TRIPS    &    \textbf{20.89}  &     \textbf{343.05}       &   \textbf{~11ms}      \\  \bottomrule[2.0pt]
\end{tabular}
\centering
\caption{The comparison of the efficiency of different models. Here, we report FLOPs, throughput and latency. The FLOPs results of the baselines come from  \cite{Kim2021ViLTVT}. Since FLOPs are proportional to input size, for a fair comparison, we keep same the input size with \cite{Kim2021ViLTVT} which is 197 for image patches length and 40 for text tokens length. We keep the same setting when calculating throughput and latency.}
\label{table_efficient}
\end{table}

\begin{table*}[h]
 \setlength{\tabcolsep}{2.0mm}{
\begin{tabular}{cccccccc}
\toprule[2.0pt]
 Locations & Keep rates    & Overall Keep rate & VQA test-dev & NLVR dev & FLOPs (G) & Throughput \\ \hline
 -                                                                  & -            & 100\%    &   76.12 & 82.35    &  76.03       & 79.32  \\ \hline
{[}2{]}                                                            & 50\%             & 50\% &   75.60  & 81.26   &   42.17    &  161.26 \\
{[}10{]}                                                           &50\%              &  50\% &  76.19  & 81.84  &   63.84   & 96.66  \\ \hline
 {[}2,4{]}                                                          & 50\%     &   \%25 & 74.21 & 79.93  &  \textbf{28.00}    &  \textbf{238.41} \\
{[}4,8{]}                                                          & 50\%    &   \%25 &  74.93 & 80.65&   38.82  &  165.30 \\
{[}5,10{]}                                                          & 50\%    &   \%25 & 75.29& 81.17 &  44.22  & 143.37 \\
{[}6,12{]}                                                        & 50\%     &   \%25 &  75.48& 81.04 &   49.63  & 126.46  \\\hline
{[}2,4{]}                                                          & 70\%    &  \%49 &  74.87 & 80.45&   43.96   &  153.92\\
{[}4,8{]}                                                          & 70\%     &   \%49 &  75.94& 81.96 &   51.72  &  125.55 \\
{[}5,10{]}                                                          & 70\%     &   \%49 & 76.23 & 82.35 &  55.60  &  115.01 \\
{[}6,12{]}                                                        & 70\%     &   \%49 &   \textbf{76.24}& \textbf{82.48} &  59.48  &  106.07 \\ \hline
{[}2,6,10{]}                                                        & 70\%  &   \%34 & 74.92 & 80.61 &   42.66  & 156.13\\
{[}3,6,9{]}                                                        & 70\%  &   \%34 &  75.09 & 80.75 & 43.49 &   151.40 \\
{[}4,8,12{]}                                                       & 70\%  &   \%34 &   75.23& 81.12 & 49.74  & 129.81 \\ \bottomrule[2.0pt]
\end{tabular}}
\centering
\caption{Results of pre-training and finetuning TRIPS with different locations and keep rates. we report the text-dev score results of VQA, FLOPs and Throughput. In this table, we set the input image size to 384 $\times$ 384 and the length of input text is 40. The throughput (image-text/s) is measured on an NVIDIA V100 GPU using the largest possible batch size for our model. }
\label{table4}
\end{table*}

\section{Experiment Results}

\subsection{Main Result}
We evaluate our model TRIPS on three widely explored vision-language downstream tasks: Visual Question Answering (VQA), Cross-modal Retrieval, and Natural Language for Visual Reasoning (NLVR). For the proposed model TRIPS, we take the 5-th and 10-th as the patch-selection layer in ViT encoder and set the keep rate of each layer to 70\%, achieving the desired trade-off between the downstream task performance and the model inference speed. Details of the datasets and fine-tuning hyperparameters are in Appendix \ref{sec:downsteam}. Details of the comparison methods are in Appendix \ref{sec:compare}.

\subsubsection{Visual Question Answering}

The VQA task \cite{Agrawal2015VQAVQ} requires the model to answer natural language questions given an image.  We follow \cite{Li2021AlignBF} and consider VQA as an answer generation problem. We report test-dev and test-std scores by submitting our results to the evaluation server\footnote{https://eval.ai/web/challenges/challenge-page/830/overview} in Table \ref{table1}. Compared with the VLP baselines, TRIPS can improve the performance on the VQA task. The results demonstrate the effectiveness of TRIPS.

\subsubsection{Natural Language for Visual Reasoning}

The NLVR2 \cite{Suhr2019ACF} task requires the model to predict whether a sentence describes a pair of images which is a binary classification task. 
 We follow \cite{Li2021AlignBF} and use two cross-attention layers to process the two input images, and their outputs are merged and fed to a Feed Forward Network (FFN). An MLP classifier is then applied to the output embedding of the text [CLS] token.  
As shown in Table~\ref{table1}, TRIPS has a better performance than existing VLP methods.

\subsubsection{Image-Text Retrieval}

We conduct experiments for both image-to-text retrieval (TR) and text-to-image retrieval (IR) on MSCOCO \cite{Lin2014MicrosoftCC} and Flickr30K \cite{Plummer2015Flickr30kEC} datasets. We jointly optimize the ITC loss and the ITM loss during fine-tuning.
The results are reported in Table \ref{table2}. As shown in Table \ref{table2}, the experimental results show that our model gets comparable performance with other VLP baselines.

\subsection{Efficiency of TRIPS}

To investigate the efficiency of Text-Relevant Image Patch Selection, we first compare the computational complexity of different models. We report the Floating Point Operations Per second (FLOPs), a widely used evaluation metric for model computational complexity. Besides, to evaluate the computational speed of our model, we compare the throughput and latency of different models. We use a Xeon Platinum 8163 CPU and an NVIDIA V100 GPU to calculate the latency and throughput. As shown in Table \ref{table_efficient}, TRIPS not only has the lowest computational complexity (e.g., 20.89 of FLOPs) but also the fastest computational speed (e.g., 343.05 of Throughput and 11ms of Latency).

\subsection{ The Impact of Location and Keep Rate }
To validate the impact of patch-selection layer location and keep rate on the efficiency and effectiveness of the model, we train TRIPS with different patch-selection locations and numbers of selected tokens. The results in Table \ref{table4} demonstrate two findings. First, moving the patch-selection layer into shallower layers reduces computational complexity but deteriorates the accuracy. For example, the accuracy drops considerably with the remarkable throughput increasing when the patch-selection layer is placed before the third layer (i.e., at the second layer). A possible explanation is that the attention maps between the text [CLS] embedding and the input tokens can be unreliable during the early processing of input tokens in shallow layers. Second, too many image tokens fused in the patch-selection layer will considerably drop the downstream task performance. For example, if we take the 2-nd and 4-th layers in ViT as the patch-selection layer and set the keep rate to 50\%, the performance will decrease to 74.21 on the VQA task, compared to 76.12 of the model without the patch-selection layer. 

\begin{table*}[h]
\centering
\setlength{\tabcolsep}{2.6mm}{
\begin{tabular}{@{}ccccccc@{}}
\toprule[2.0pt]
 Selection Layer         & Keep rate     & image size     & VQA test-dev& NLVR dev  & FLOPs(G) & Troughout \\ \midrule
 - & - & $384 \times 384$ &  76.12 & 82.35  & 76.03   & 79.32 \\ \hline
 {[}5,10 {]}            & 70\% & $224\times224$ &     75.23 & 80.83   &\textbf{ 20.89} &  \textbf{ 343.05}   \\
 {[}5,10 {]}            & 70\% & $256\times256$ &      75.84   & 81.24  &    26.61  &  258.03     \\
 {[}5,10 {]}            & 70\% & $304\times304$ &    76.13      & 81.72 &    36.62   &  189.07\\
 {[}5,10 {]}            & 70\% & $384\times384$ &      76.23   & 82.35  &    55.60 &  115.01    \\
  {[}5,10 {]}            & 70\% & $456\times456$ &     \textbf{ 76.54}   &\textbf{ 83.02}  &  74.83  &  81.0   \\\bottomrule[2.0pt]
\end{tabular}}
\centering

\caption{Results of TRIPS finetuning on VQA and NLVR task with different resolution images. When calculating FLOPs, the input length of the text is kept at 40 and the settings for calculating throughput are the same as Table\ref{table4}.}
\label{table5}

\end{table*}

\begin{table*}[h]
\setlength{\tabcolsep}{1.5mm}
\begin{tabular}{ccccccccc}
\toprule[2.0pt]
\multirow{2}{*}{Models} & \multirow{2}{*}{Selection Layer} &\multirow{2}{*}{Keep rate} &\multirow{2}{*}{FLOPs(G)}&\multirow{2}{*}{Throughput} & \multicolumn{2}{c}{VQA} & \multicolumn{2}{c}{NLVR} \\
                        &               &  &  &            & Test-dev   & Test-std   & dev        & Test-P      \\ \hline
                        
TRIPS-S        & [5,10] & 70\%        &    59.42   &    \textbf{216.74}      & 71.48      & 71.52      & 75.89       &  76.47    \\
 -$w/o$ PS     & - &     -     &   104.42         &    135.25     & 71.26      & 71.29  &  75.18   &   76.23     \\ \hline
TRIPS       &[5,10] & 70\%     & \textbf{55.60}    &  115.01      &  \textbf{76.23}     & \textbf{76.48}     & \textbf{  82.35}   & \textbf{83.34}        \\
 -$w/o$ PS  & - &    -      &  76.03        &  79.32  &  76.12    &  76.32      &    82.21   &    83.11      \\ \bottomrule[2.0pt]
\end{tabular}
\centering
\caption{Ablation results of TRIPS and TRIPS-S on VQA test set and NLVR2.$w/o$ PS indicates that we remove the patch-selection module in ViT. The setting for calculating FLOPs and throughput is the same as Table\ref{table4}.}
\label{table3}

\end{table*}

\subsection{Finetuning on Higher Resolution Images}

We can control the computational cost by fusing different numbers of inattentive tokens. Therefore, we finetune TRIPS on the VQA and NLVR tasks, which take images with varying resolutions as input. We report the results in Table \ref{table5}. The experimental results show that by increasing the input image resolution, we can facilitate the model by taking more image tokens to gain better performance. For example,  by finetuning TRIPS with the images of 456$\times$456, we can achieve the score of 76.54 on VQA, outperforming the baseline finetuned with images of 384$\times$384 yet keeping similar computational complexity. 

\subsection{Effectiveness of Text-Relevant Image Patch Selection }
To verify the effectiveness of Text-Relevant Image Patch Selection, we first implement the single-stream model TRIPS-S as we present in subsection \ref{sect2}. Then, we examine the downstream task performance, computational complexity, and inference speed of TRIPS and TRIPS-S (both with and without Text-Relevant Image Patch Selection). The results are shown in Table \ref{table3}, and we find that for both TRIPS and TRIPS-S, we can see a consistent improvement in the inference speed and downstream task performance by incorporating the text-relevant image patch selection mechanism. These results suggest that the proposed image patch selection mechanism is not only efficient but also effective. Notably,  compared with the dual-stream model TRIPS, TRIPS-S is faster in inference due to the parameter efficiency of the single-stream model. However, its performance lags behind state-of-the-art performance on downstream VL tasks.

\begin{table}[h]
\setlength{\tabcolsep}{1.5mm}{
\begin{tabular}{@{}lcccc@{}}
\toprule[2.0pt]
model    & VQA & FLOPs(G) & Throughput \\ \midrule
TRIPS     &  76.23   & 57.20   &  111.83        \\  
  -$w/o$ ITF &  75.92 &    57.15  &     112.04     \\ 
  -$w/o$ TD-ATT &   75.23      &  56.4     &     117.21    \\ \bottomrule[2.0pt]
\end{tabular}}
\caption{The result of ablations. We finetune TRIPS on VQA and report test-dev results. The setting for calculating FLOPs and throughput is the same as Table \ref{table4}. The same with the settings of TRIPS we present in the main results, we select the 5-th and 10-th as the patch selection layer, and each layer will keep 70\% image tokens. }
\label{table6}

\end{table}
\subsection{Ablation Study}
 We also perform ablation studies to investigate the effects of inattentive image token fusing and Text-aware Dynamic Attention. In Table \ref{table6}, $w/o$ ITF indicates the inattentive tokens are directly discarded without fusing.  As shown in Table \ref{table6}, fusing inattentive tokens outperforms the model without inattentive tokens. Although the improvement is small, there is no additional computational overhead introduced. We also verify the impacts of Text-aware Dynamic Attention (TD-ATT). Specifically, $w/o$ TD-ATT indicates that we remove the TD-ATT in the patch-selection layer and select the image tokens based on the image [CLS] token. As shown in Table \ref{table6}, selecting the image patch tokens with the image [CLS] token without considering the linguistic context will degrade the model's performance. This result supports our motivation that directly removing patch tokens based on image [CLS] without incorporating the text knowledge is unsuitable for VLP models.
 
\subsection{Visualization}

The proposed TRIPS accelerates VLP by a novel patch selection module that selects the text-consistent image tokens in the vision backbone and preserves the attentive image tokens. To further investigate the interpretability of our model, we visualize the procedure of text-relevant image path selection in Figure \ref{fig3}. It can be seen that as the network deepens, the inattentive tokens are gradually removed or fused, while the text-relevant tokens are selected and preserved. 
Besides, we present more visualization results in Figure~\ref{fig4} to show the effectiveness of the text-relevant image patch selection module. We take different text words as the input and visualize the text-aware image patches selected by the text-relevant image patch selection module. As shown in Figure~\ref{fig4}, the selected image patches are highly relevant to the query texts and thus enable our model to make a correct prediction.

\section{Conclusion}
We have presented TRIPS, an efficient VLP model with \textbf{T}ext-\textbf{R}elevant \textbf{I}mage \textbf{P}atch \textbf{S}election to progressively reduce the redundant image tokens with text guidance. TRIPS introduces a novel patch selection module to select the text-consistent image tokens in the vision backbone, which preserve the attentive image tokens with text guidance and fuses the inattentive tokens into one token by dynamically computing text-dependent visual attention in an end-to-end way. The experiment shows our method not only decreases the computation cost of VLP but also improves the efficiency of cross-modal fusion due to the reduction of visual sequences, while keeping or even improving the performance of downstream image-text tasks.

\section{Limitations}
Despite the effectiveness and efficiency of TRIPS across a wide range of downstream image-text tasks, our model still has several limitations. First, in our current settings, we pre-train TRIPS with only 4M image-text pairs. It is unclear how well the performance will be if we pre-train TRIPS on a larger pre-training dataset with other available data types, such as text-only, image-only data, and some labeled data. Second, we provide a novel perspective for the efficient training and inference of VLP models, which reduces the visual sequence progressively with a text-guided patch-selection layer in the visual backbone. In our method, the number of selected text-aware image patches in each patch-selection layer is fixed, and there can be a more ingenious technical design that can dynamically select different numbers of image patches.  

\section{Acknowledgement}
This research was supported by the National Key Research and Development Program of China(No. 2021YFC3340101).
\bibliography{anthology,custom}
\bibliographystyle{acl_natbib}

\appendix
\begin{figure*}[h]
\setlength{\belowcaptionskip}{-0.15cm}
\centering
\includegraphics[width=0.6\textwidth]{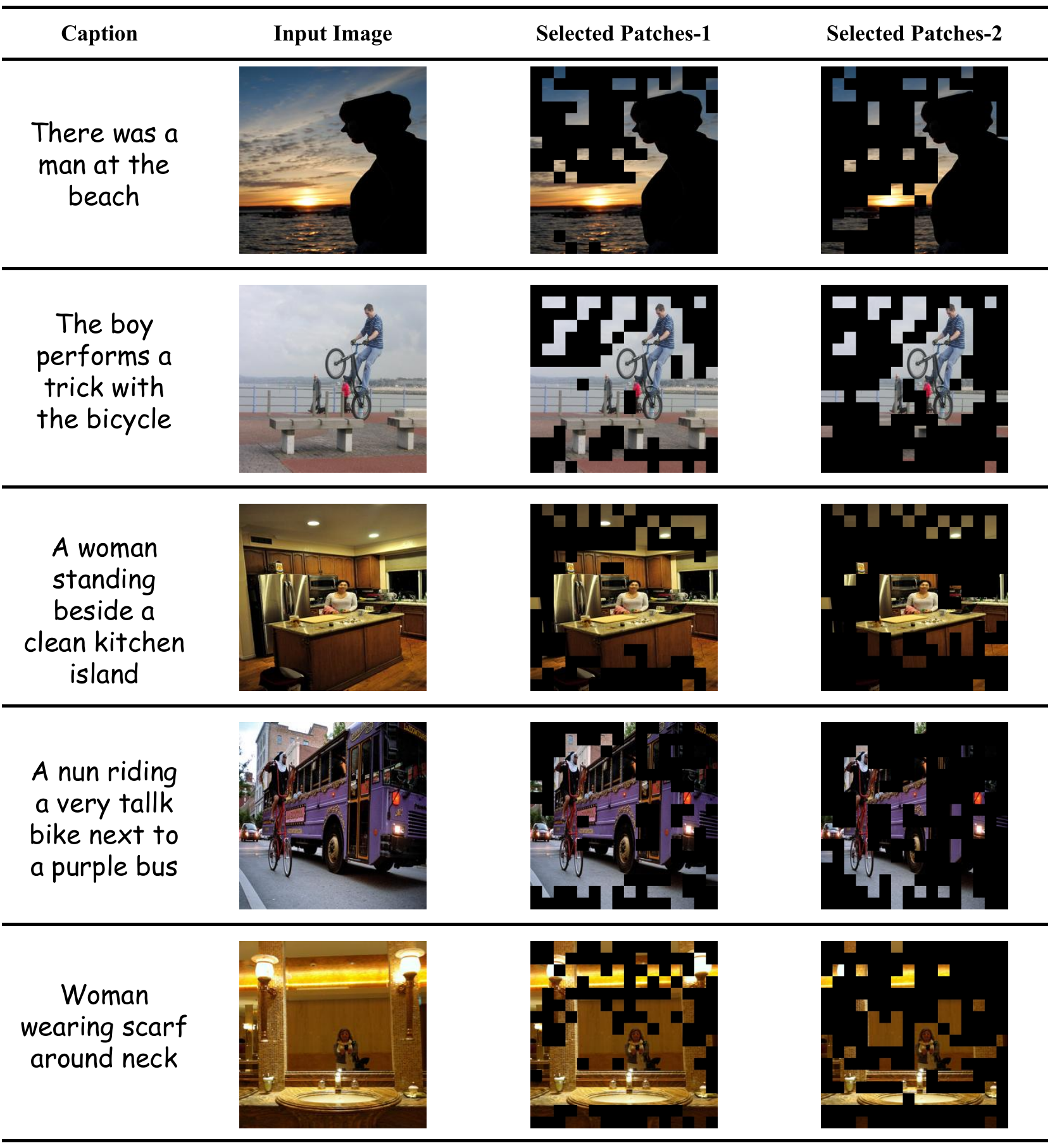}
\caption{The visualization of the selected text-aware image patches in different selection layers. We set the 5-th and 10-th layers in the vision backbone as the patch-selection layer and we keep 70\% image patches in each layer.  }
\label{fig3}
\centering
\end{figure*}
 \begin{figure*}[htpb]
\setlength{\belowcaptionskip}{-0.15cm}
\centering
\includegraphics[width=0.9\textwidth]{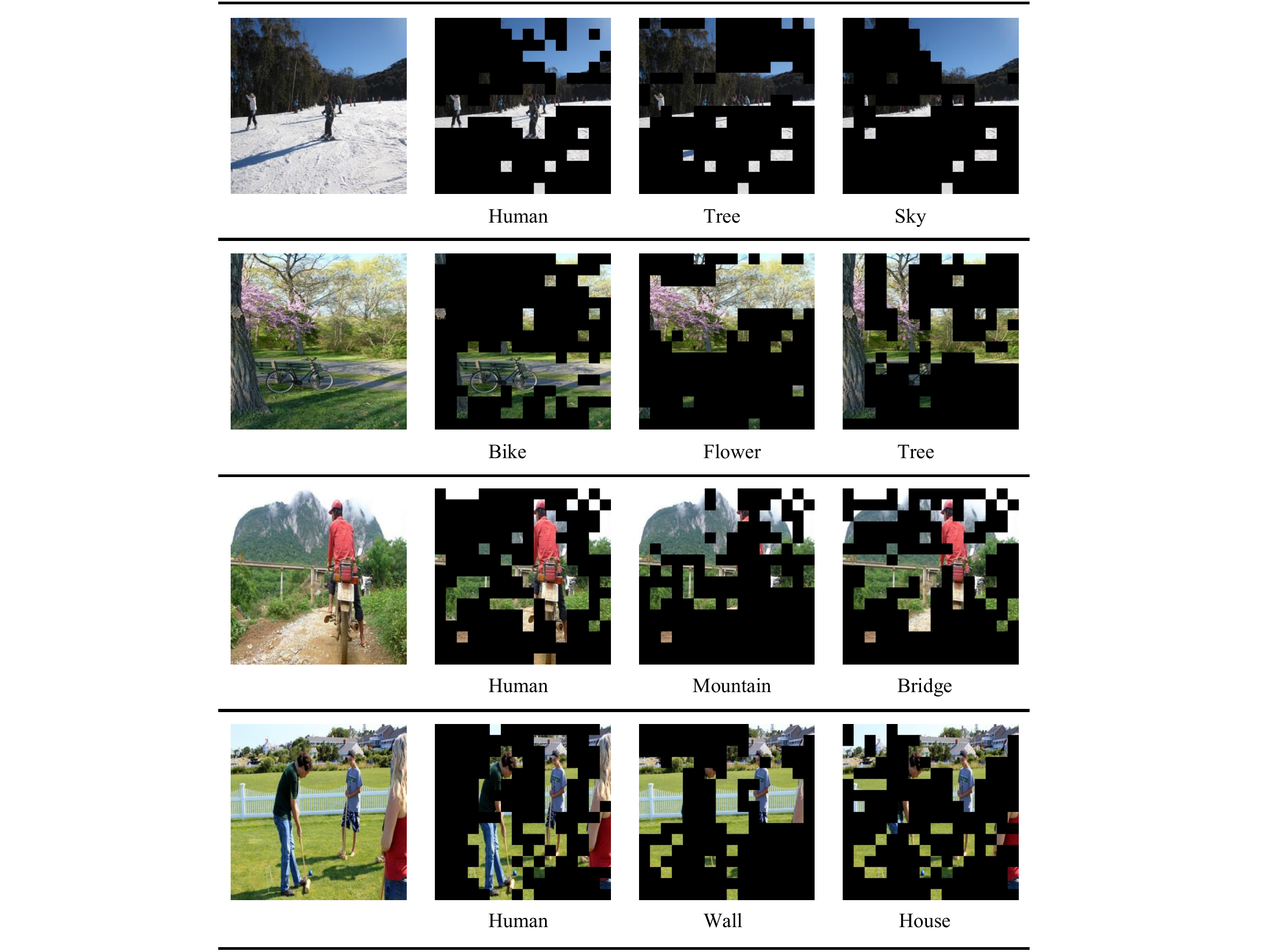}
\caption{The visualization of the selected text-aware image patches with different text words. We set the 5-th and 10-th layers in the vision backbone as the patch-selection layer and we keep 70\% image patches in each layer. We visualize the selected text-aware image patches output by the 10-th layer. }
\label{fig4}
\end{figure*}

\clearpage

\section{Comparison Methods}
\label{sec:compare}
\vspace{-3cm}
\noindent \textbf{LXMERT}~\cite{Tan2019LXMERTLC}: is the first two-stream region-based VLP model, which consists of an object relationship encoder, a language encoder and a cross-modality encoder.  

\noindent \textbf{E2E-VLP}~\cite{Xu2021E2EVLPEV}: proposes the first end-to-end VLP method for both V+L understanding and generation, with a unified Transformer encoder-decoder architecture.

\noindent \textbf{VILT}~\cite{Kim2021ViLTVT}:  adopts linear projection and word embedding as the visual and textual encoders, and uses the visual transformer as the cross-modal encoder to align and fuse the features of both modalities in an end-to-end manner.

\noindent \textbf{OSCAR}~\cite{Li2020OscarOA}: proposes to use object tags detected in images as anchor points to the learning of cross-modal alignments.

\noindent \textbf{VinVL}~\cite{zhang2021vinvl}: pre-trains a large-scale object-attribute detection model with much larger amounts of supervised data to extract better region-based visual features. 

\noindent \textbf{ALBEF}~\cite{Li2021AlignBF}: adopts a contrastive loss to align the image and text representations, then fuses them through cross-modal attention in an end-to-end manner.

\noindent \textbf{UNITER}~\cite{Chen2019UNITERLU}: proposes a new word-region alignment pre-training task via the use of optimal transport to help fine-grained alignment between words and image regions. 

\noindent \textbf{ViLBERT}~\cite{Lu2019ViLBERTPT}: proposes one of the first work that extend the BERT architecture to a multi-modal two-stream region-based VLP model.

\section{Downstream Task Details}
\label{sec:downsteam}

We evaluate TRIPS on the three downstream vision-language tasks. The hyperparameters that we use for finetuning on the downstream  tasks are listed in Table \ref{table:finetune-hyper}. Following ~\citep{li2021align}, all tasks adopt RandAugment, AdamW optimizer with a weight decay of 0.05 and a cosine learning rate schedule. Next we introduce the dataset settings in detail.
\begin{table}[htbp]
\setlength\tabcolsep{1pt}
\centering
\begin{tabular}{l|ccc}
\toprule
Task  &  LR (ViT-B/BERT$_{base}$) & batch size & epochs  \\
\midrule
VQA & 2e-5/5e-6 & 1024 &  8 \\
Retrieval & 1e-5/2e-6 & 256& 5 \\
NLVR2 & 5e-5/5e-6 & 256 & 15 \\
\bottomrule
\end{tabular} 
\caption{Finetuning hyperparameters for downstream tasks.}
\label{table:finetune-hyper}
\end{table}

\paragraph{VQA.} The VQA task ~\cite{Agrawal2015VQAVQ} requires the model to answer natural language questions given an image. We conduct experiment on the VQA2.0 dataset ~\citep{Agrawal2015VQAVQ}, which contains 83k/41k/81k images for training/validation/test. Following ~\citep{li2021align}, we use both training and validation splits for training, and incorporate additional training data from Visual Genome~\citep{Suhr2019ACF}. 

\paragraph{Image-Text Retrieval.} We conduct experiments for both image-to-text retrieval (TR) and text-to-image retrieval (IR) on COCO ~\cite{Lin2014MicrosoftCC} and Flickr30K ~\cite{Plummer2015Flickr30kEC} datasets. We take the widely-used Karpathy split ~\citep{karpathy2015deep} for both COCO and Flickr30K. COCO contains 113k/5k/5k images for train/validation/test, and Flickr30K contains 29k/1k/1k images for train/validation/test. 

\paragraph{NLVR2.} The NLVR2~\cite{Suhr2019ACF} task requires the model to predict whether a sentence. We conduct experiments following the original train/val/test split in ~\cite{Suhr2019ACF}. 


\end{document}